\documentclass{article}
\usepackage{ijcai19}
\usepackage{times}
\usepackage{soul}
\usepackage{url}
\usepackage[hidelinks]{hyperref}
\usepackage[utf8]{inputenc}
\usepackage[small]{caption}
\usepackage{graphicx}
\usepackage{amsmath}
\usepackage{amsfonts}
\usepackage{booktabs}
\usepackage{algorithm}
\usepackage{algorithmic}
\usepackage{fancyhdr}

\title{Optimizing Hyperparameters in CNNs using Bilevel Programming\\ for Time Series Data}

\author{
Taniya Seth\and
Pranab K. Muhuri\\\
\affiliations
Department of Computer Science, South Asian University, New Delhi, India\\
\emails
taniya.seth@students.sau.ac.in,
pranabmuhuri@cs.sau.ac.in
}

\begin{document}

\maketitle

\begin{abstract}
  Hyperparameter optimization has remained a central topic within the machine learning community due to its ability to produce state-of-the-art results. With the recent interest growing in the usage of CNNs for time series prediction, we propose the notion of optimizing Hyperparameters in CNNs for the purpose of time series prediction. In this position paper, we give away the idea of modeling the concerned hyperparameter optimization problem using bilevel programming.
\end{abstract}

\section{Introduction}

Training a machine to perform humanely tasks such as image recognition and data prediction, involves preparing a {\it good enough} model that learns the given data. This model predominantly involves a training algorithm that is responsible for this learning task. Furthermore, the job of this training algorithm is to develop a function, which in essence minimizes a loss on some data samples (a subset of the ground truth data) introduced to it. This trained model is then applied to the test data (and out-of-sample subset of the ground truth data), on which the model is evaluated based on another loss.

Who evaluates the performance of this model? Who decides exactly {\it how much} is {\it good enough}?

Literature on machine learning has blessed us with answers to such questions while actually producing {\it good models} to make machines perform various tasks.

The answer to the above question is that, the evaluation of the training algorithm in the model is done using a training loss, which identifies the difference between the actual state of the model’s learning and the training data it is provided with to learn. The function mentioned earlier in this section minimizes this difference between the learnt data and the training data. This is, with respect some parameters of the model, say {$\theta$}. The training algorithm gradually learns these parameters during the concerned process, model weights, for example.

However, a model also includes the hyperparameters, {$\lambda$} in the scene, which ~\cite{cite01} refer to as the “bells and whistles” of a training algorithm. In practice, hyperparameters are chosen first, which is then followed by the development of the training algorithm. Due to their importance and influence on the training, these hyperparameters require expert intervention to be chosen.

When optimized values of hyperparameters are supplied to the training algorithm, it learns well from the training data, while additionally performing well on the out-of-sample test data. The performance of the model on the test data is evaluated based on a validation loss, which must be minimized for the model to generalize well.

This discussion defines the necessity of hyperparameter optimization (HO) within machine learning models. This problem has been studied for a long time. ~\cite{cite02} utilized various approaches such as the sequential model based approach, Gaussian process approach, tree-structure Parzen estimator approach etc for optimizing the estimated improvement criteria. Random search was subsequently studied for HO in ~\cite{cite01}. Later, ~\cite{cite03} introduced Auto-WEKA for the combined selection and HO in classification algorithms. ~\cite{cite04} put forward an empirical study to deal with Bayesian optimization for hyperparameters. Most importantly, gradient-based HO was discussed in ~\cite{cite05}, wherein exact gradients of hyperparameters were computed by chaining their derivatives backwards in the training procedure through reversible learning. Other works on HO include ~\cite{cite06} and ~\cite{cite07}.

Having discussed the problem of HO above, one can notice the dual structure that the problem encompasses. In other words, performance of a machine learning model is optimized based on the training and validation losses. This optimization is subject to the values chosen for hyperparameters, {$\lambda$} of the model. This can be stated as the following: the validation loss of a model is minimized with respect to minimized training loss, for the model which is parameterized by the hyperparameters.

Such a dual structure is noticed in multiple real-life situations, which can be modeled using the bilevel programming strategy ~\cite{cite08}. Solving these problems follow a leader-follower approach, inspired from the game theory ~\cite{cite09}. Within these problems, the solution space of the objective function (OF) of the leader is constrained by that of the follower problem. Hence, a proper solution is sought that satisfies both the leader’s and follower’s solution space while optimizing their individual objectives.

Recently, the idea of HO using bilevel programming was proposed in ~\cite{cite10}. Franceschi and co-authors developed a bilevel optimization framework for HO. Upon formulating the bilevel for HO, they observed that it is difficult to obtain a solution to the bilevel model, especially when {$\lambda$} is a real-valued vector of hyperparameters. To overcome this, the exact problem of the bilevel model was approximated and later proven to guarantee solutions.

In the literature so far, the problem of HO has been dealt with mostly for the cases of images. In today’s world, time series is available in abundance. From stock market to daily average temperatures, human activity data and now most importantly COVID-19 data, everything is available as a time series. Leveraging such series for either classification or prediction is crucial. Convolutional neural networks (CNN) have been utilized for both classification and prediction purposes on time series data.

~\cite{cite11} time series data utilizing multiple channels deep CNNs with special attention to exploration of feature learning techniques. In ~\cite{cite12}, classification of the human activity recognition (HAR) is done using deep CNNs, whereas in ~\cite{cite13}, time series classification is done using multiscale CNNs. Other recent works on time series forecast and classification using CNNs in-clude ~\cite{cite14} and ~\cite{cite15}.

Keeping an eye on the relevance of time series prediction in today’s world, one can observe that the literature lacks works where a machine learning model has been optimized for performance on time-series data.

Hence, in this position paper, we propose the idea of utiliz-ing bilevel programming for HO within CNNs for time series prediction. We first introduce the bilevel framework to model the overall performance of the machine learning model in terms of the training and validation loss. This is done in Section ~\ref{section2}. Subsequently, in the same section, we revisit the approximation strategy for the bilevel framework of HO, along with the gradient-based approach to solve the problem. In Section ~\ref{section3}, we introduce our proposed framework of utilizing bilevel programming for HO in CNNs for time series prediction. We conclude the position paper in Section ~\ref{section4}.

\section{Preliminary Knowledge}
\label{section2}

\subsection{Bilevel programming framework}

In this section, we revisit the structure of the bilevel programming framework for a machine learning model. As specified in ~\cite{cite10}, bilevel programming problems of the following forms are considered:

\begin{equation}
\label{equation1}
    \min\{f(\lambda):\Lambda\in\lambda\}
\end{equation}
where,

\begin{equation}
\label{equation2}
    f(\lambda)=\inf\{E(w_{\lambda},\lambda):w_{\lambda}\in\arg\min_{u\in\mathbb{R}^{d}}L_{\lambda}(u)\}
\end{equation}

In the above equations, {$f:\Lambda\rightarrow{\mathbb{R}}$} is defined at {$\lambda\in\Lambda$}. {$E:\mathbb{R}^{d}\times\Lambda\rightarrow{\mathbb{R}}$} is the leader objective. Also, {$\forall\lambda\in\Lambda$}, {$L_{\lambda}:\mathbb{R}^{d}\rightarrow{\mathbb{R}}$} is the follower objective given that {$L_{\lambda}:\lambda\in\Lambda$} is the class of OFs parameterized by {$\lambda$}.

\subsection{Bilevel programming framework for HO}

As mentioned earlier, the validation error is sought to be minimized for a machine learning model. Let the model be denoted as {$g_w:X\rightarrow{Y}$} Let it be parameterized by the vector {$w$}, with respect to one vector of hyperparameters {$\lambda$}. For a predefined loss function {$l$}, the leader and follower objectives can be given as follows:

\begin{equation}
\label{equation3}
    E(w,\lambda)=\Sigma_{(x,y)\in{D_{validation}}}l(g_{w}(w),y)
\end{equation}

\begin{equation}
\label{equation4}
    L_{\lambda}(w)=\Sigma_{(x,y)\in{D_{train}}}l(g_{w}(w),y)+penalty
\end{equation}

Here, {$D_{validation}$} is the validation data presented to {$g_w$}, for evaluation after it has been trained on {$D_{train}$}. The penalty term can be implemented as a regularizer for the network model to improve the performance.

\subsection{Gradient based approach to solve bilevel optimization for HO}

\cite{cite10}, specified an approximation of the bilevel problem given in (\ref{equation1}) and (\ref{equation2}). It is given as follows:

\begin{equation}
\label{equation5}
    \min_{\lambda}f_{T}(\lambda)=E(w_{T,\lambda},\lambda)
\end{equation}

\begin{equation}
\label{equation6}
    w_{0,\lambda}=\phi_{0}(\lambda), w_{t,\lambda}=\phi_{t}(w_{t-1,\lambda},\lambda), t\in{[T]}
\end{equation}

In the above equations, {$[T]$} is a predefined positive integer such that {$[T]=\{1,\dots,T\}$}, {$\phi_{0}:\mathbb{R}^{m}\rightarrow{\mathbb{R}^{d}}$} is a smooth initialization dynamic, and {$\forall{t}\in[T]$}, {$\mathbb{R}^{d}\times\mathbb{R}^{m}\rightarrow{\mathbb{R}^d}$}is a smooth mapping the operation of an optimization algorithm at the {$t^{th}$} step. The optimization dynamic {$\phi$} is implemented using the gradient descent optimization algorithm.

In ~\cite{cite10}, certain assumptions are chosen to reduce the bilevel framework given in (\ref{equation1})-(\ref{equation2}), to prove the existence of solutions of the reduced problem and also the existence of the convergence of approximate problems to the reduced problem. They are omitted from this position paper for simplicity.

\section{HO using bilevel programming within CNNs}
\label{section3}

We discuss our proposed idea in this section.

We first define our CNN model for classification purposes. For our time series data, we utilize 1D convolutional layers, which are fit for situations dealing with time series information.

\begin{figure}[t]
\includegraphics[width=8cm]{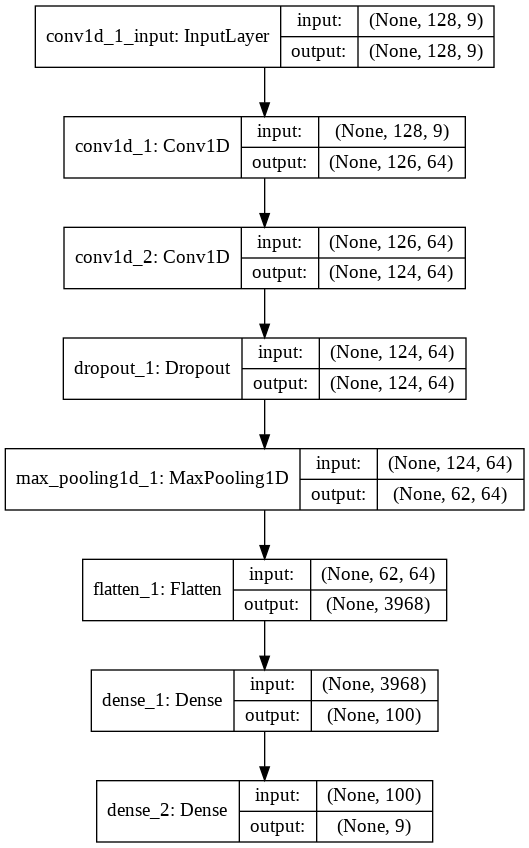}
\caption{Network structure for time series prediction using a deep CNN architecture}
\label{Fig1}
\end{figure}

For explanation, we utilize a time series dataset with 128 time steps and 9 features of data. Our deep CNN model for this time series data begins with an input layer, followed by two 1D convolutional layers each encompassing 64 filters with a filter size of 3. Both layers have the ReLU activation applied. These are followed by a dropout layer with a 50\% dropout rate, followed by a max pooling layer. The output from the max pooling layer is then flattened and forwarded to a dense layer with 100 connections and the ReLU activation, followed by a final dense layer with 9 output units and the softplus activation. The model structure is depicted in Fig. ~\ref{Fig1}.

For this model and data, we consider the example weights ({$w$}) and the learning rate ({$lr$}) of the neurons as the hyperparameters to be optimized. The metric to be minimized is given by the model is the Mean Squared Error (MSE), while the optimizer utilized is the Adam optimizer.

With this scenario defined, the bilevel programming framework for our CNN model for the time series data is described below. For the following, {$\lambda=\{w,lr\}$} and {$T=200$}.

\begin{equation}
\label{equation7}
   \min_{\lambda}f_{T}(\lambda)=Adam(w_{T,\lambda},\lambda)
\end{equation}
and,

\begin{equation}
\label{equation8}
\begin{split}
   \phi_{0}(\lambda)=\{w_{0}=[0],lr=0.01\}, \\ \phi_{t}(w_{t-1,\lambda},\lambda)=w_{t}-\eta_{t}\nabla{L_{\lambda}}, t\in{[T]}
\end{split}
\end{equation}

The follower level optimizer is defined by the gradient descent optimizer as given in ~\cite{cite10}. This follower level optimizer is defined for the hyperparameter, {$lr$}. While the {$w$} is the minimizer for the problems in (~\ref{equation7})-(~\ref{equation8}).

We believe that solving this bilevel problem to obtain the optimized value of {$lr$} with respect to the minimizer {$w$}, shall produce state-of-the-art results in terms of MSE.

We plan to implement this scenario in on a machine with the following specifications: Intel Core 140 i3-6100 CPU with 12 GB of RAM and Windows 10 OS. The GPU employed is the NVIDIA GeForce 141 GTX 1660 Super.

\section{Conclusions and Future Work}
\label{section4}

In this position paper, we have introduced the idea of using bilevel programming for HO within CNNs for time series data. Since the literature on HO for time series prediction or classification tasks is scarce, we believe that the idea presented here will mark a good start in the research in this direction.

We utilized a deep CNN architecture to define the model for the purpose of time series prediction. Based on this, we defined a framework for the bilevel programming problem that must be solved to obtain the better results than most of the existing models.

Our subsequent plans are to implement the scenario introduced within this position paper. Within this implementation, we shall perform a sensitivity analysis on different values of {$T$}, to obtain varied results. We also plan to compare the impact of HO using bilevel programming within the prediction and classification tasks on time series data. We plan to perform our experiments, on the human activity recognition (HAR) data to observe the results.

\bibliographystyle{named}
\bibliography{bibliography}

\end{document}